\documentclass[runningheads]{llncs}
\usepackage{graphicx}
\begin{document}
\title{MeDiaQA: A Question Answering Dataset on Medical Dialogues}
%
%
\author{Huqun Suri \and Qi Zhang\and Wenhua Huo \and Yan Liu \and Chunsheng Guan}
\authorrunning{Suri et al.}
%
\institute{Institute of Science and Technology, Taikang Insurance Group\\ 
\email{taikang-ai@taikanglife.com}\\
}
\maketitle              
\begin{abstract}
In this paper, we introduce MeDiaQA, a novel question answering(QA) dataset, which constructed on real online \textbf{Me}dical \textbf{Dia}logues. It contains 22k multiple-choice questions annotated by human for over 11k dialogues with 120k utterances between patients and doctors, covering 150 specialties of diseases, which are collected from haodf.com and dxy.com. MeDiaQA is the first QA dataset where reasoning over medical dialogues, especially their quantitative contents. The dataset has the potential to test the computing, reasoning and understanding ability of models across multi-turn dialogues, which is challenging compared with the existing datasets. To address the challenges, we design MeDia-BERT, and it achieves 64.3$\%$ accuracy, while human performance of 93$\%$ accuracy, which indicates that there still remains a large room for improvement.

\end{abstract}
\section{Introduction}
Recently, telemedicine is becoming an important complement to traditional face-to-face medicine practiced in hospitals, since it can delivering patient care remotely, there are many advantages including increasing access to medical care especially for people living in communities that are in shortage of hospitals, and reducing healthcare costs. The studies in (Pande and Morris, 2015) and (Wootton et al., 2017) also show telemedicine can improve the satisfaction level of patients. And the fact that most of the telemedicine services are based on conversations between doctors and patients in text suggests that understanding user-generated medical dialogues is a significant task.

The research of Question answering (QA), to build a system to understand provided reading materials and answer corresponding questions, has attracted many researchers’ attention in past years. And many various QA datasets have been built to study how to make a QA system to understand a specific passage, the common sense knowledge and etc. However, although these datasets are based on different domains, they share virtually the same design pattern, most questions in these datasets could be answered by extracting a few relevant sentences so that the model only needs to search and determine a small set of supporting evidences, and the order of these supporting materials does not affect the final answer, which means these questions can be solved purely by considering a fixed document. But when it comes to the real-world questions, a good model needs more to give answers, including external professional knowledge, the ability of doing mathematics calculations and recognizing different characters in text, and so on. For instance, the question “does the patient have to receive operation?” could not be answered before analyzing the whole conversation since the answer may continuously change during the discussion between doctor and patient on the condition, and sometime there even have no conclusion appeared in the dialogue, which means the answer will be given both on the condition and external medical knowledge. Another question “how many tablets the patient need to take in a week?” need the model to do mathematics calculation since the amounts provided in the dialogue maybe are for a day or a month.Therefore, with the analysis above, we think there is little work on question answering over medical dialogues, and the lack of available datasets is one of the main factor.  

Whatsmore, the questions in our real world often come with mathematics calculation, including multiplication, addition and conversion. To give correct answers, the model need to find and extract corresponding text and do calculation on the evidence. 

For all these purposes, we built a QA dataset MeDiaQA, a set of Chinese telemedicine dialogues from haodf.com and dxy.com, which are the main-stream websites for online medical consultation in China. On these websites, the patients first show their electric health records, then the doctors will discuss with the patients about their conditions and give advice. Figure 1 shows several examples in the MeDiaQA dataset. To solve the first two questions, the ability of mathematical calculation is required, while the last two needs reasoning ability.

To conclude, comparing with the existing datasets, our MeDiaQA has three distinct features. First, to answer the questions, a model need to understand medical knowledge. Second, since the dialogues are based on two characters, the model needs to track and judge the information given by different person. Third, in this dataset, to get the correct answer, the model not only need to extract corresponding evidence from the dialogue, but also do mathematics calculation.The analysis and experimental results show how this dataset can effectively examine how a QA system can handle in medical dialogue understanding and mathematical calculation which is not covered by previous datasets.

\begin{figure}
\includegraphics[width=\textwidth]{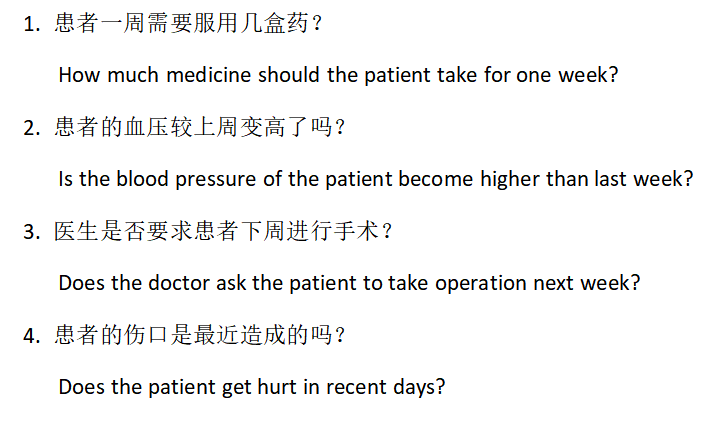}
\caption{Question Examples from the MeDiaQA dataset.} \label{fig1}
\end{figure}

The rest parts of this paper are organized as follows: In section 2, we give a brief introduction of current research on medical QA and multiple-choice QA. Section 3 describes how we constructed the dataset. And Section 4 shows statistics of the dataset and analyzes the medical term understanding and computational inference in the data. Section 5 introduces the MeDia-BERT and evaluates the results of experiments and in Section 6, we concludes the paper. 

\section{Related Works}
Machine reading comprehension (MRC) aims to answer questions by comprehending evidence from passages. This direction has recently drawn much attention due to the fast development of deep learning techniques and large-scale datasets.
And researchers propose even more challenging datasets that require QA within dialogue or conversational context (Reddy et al., 2018; Choi et al., 2018). It is worth mentioning that in almost all previously developed datasets, the passages are from Wikipedia, news articles or fiction stories, which are considered as the formal language. Yet, there is little effort on QA over informal one like medical dialogues. And according to the features of our dataset, we introduce related works in two categories.

\noindent\textbf{Medical QA:} The medical domain poses a challenge to existing approaches since the questions may be more challenging to answer. BioASQ (Tsatsaronis et al., 2012, 2015) is one of the most significant community efforts made for advancing biomedical question answering (QA) systems. SeaReader (Zhang et al., 2018) is proposed to answer questions in clinical medicine using documents extracted from publications in the medical domain. Yue et al. (2020) conduct a thorough analysis of the emrQA dataset (Pampari et al., 2018) and explore the ability of QA systems to utilize clinical domain knowledge and to generalize to unseen questions. Jin et al. (2019) introduce PubMedQA where questions are derived based on article titles and can be answered with its respective abstracts. Recently, pre-trained models have been introduced
to medical domain (Lee et al., 2020; Beltagy et al., 2019; Huang et al., 2019a). They are trained on unannotated biomedical texts such as PubMed abstracts and have been proven useful in biomedical question answering.

\noindent \textbf{Multiple-choice QA:}Datasets such as HotpotQA (Yang et al., 2018), Natural Questions (Kwiatkowski et al., 2019), ShARC (Saeidi et al., 2018) and BioASQ (Tsatsaronis et al., 2015) contain multiple-choice questions as well as other types of questions. BoolQ (Clark et al., 2019) specifically focuses on naturally occurring yes/no questions, and those questions are shown to be surprisingly difficult to answer. Recent breakthroughs of pre-trained language models like ELMo (Peters et al., 2018) and BERT (Devlin et al., 2018) show significant performance improvements on NLI tasks. In this work, we use a modified versions of them to set baseline performance on MeDiaQA. The Figure 2 shows some examples in MeDiaQA dataset, which will be discussed in the following sections.

\section{Dataset}
In this section, we mainly introduce the three-step data collection process of MeDiaQA: dialogues crawling, data processing and question-answer writing. 

\subsection{Data Crawling}
In haodf.com and dxy.com, all the dialogues between patients and doctors has a unique ID connected with their URLs. We first collect the IDs from telemedicine pages of the websites. And then we use tool to make access to all the dialogues by changing IDs in the URLs.

\subsection{Data Processing}
The raw data seems like a complete record of the process of telemedicine, including the name of doctor and hospital, the date and other information to introduce the background of patient. Since we only need to focus on the conversation itself, we remove all the irrelevant text and keep the description of condition from the patient and dialogue. And due to the data is collected randomly from the websites, the quality of the dialogues are various. To guarantee there is enough quantity of information for question writing, we only keep the dialogues which have more than 6 turns in the dataset. 

\begin{figure}
\includegraphics[width=\textwidth]{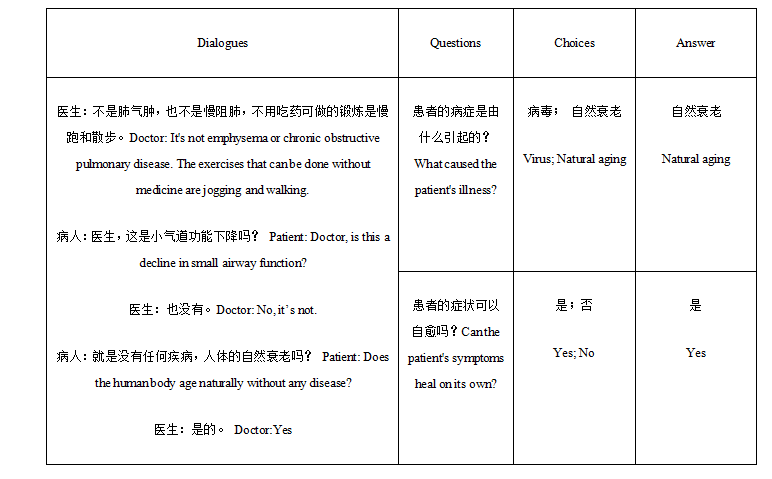}
\caption{The different types of proposed questions.} \label{fig3}
\end{figure}

\subsection{Question-Answer Writing}
We then co-work with the data labeling team to write question-answer pairs for the processed data, and for each dialogue, we write two question-answer pairs and each question has two candidate answers. To guarantee the quality and avoid trivial questions that can be simply answered by superficial text matching methods, the question writing is instructed by following items:
\begin{itemize}
    \item[*] The question should be answered by reasoning.
    \item[*] Medical background knowledge should be considered in the question.
    \item[*] No other domain questions should be asked.
    \item[*] The data used for computational questions should be located directly in the dialogues.
\end{itemize}

\section{Dataset Statistics}

\begin{table}
\caption{The details of statistics of the dataset.}\label{tab1}
\begin{center}
\begin{tabular}{|l|l|l|l|l|}
\hline \bf Element & \bf Count  & \bf Split & \bf Dialogues(Num) & \bf Questions(Num)\\ \hline
Dialogues & 11524 & Train & 8524 &17048 \\ 
Utterances in total & 123632 & Val & 1340 &2680\\
Questions in total & 23048 & Test & 1660 &3320\\
\hline
\end{tabular}
\end{center}
\end{table}

The Table 1 shows the statistics of our dataset. The MeDiaQA dataset contains 11,524 dialogues, each of which has 11 turns and 2 question-answer pairs on average. And then we analyze the questions from the dataset, according to different methods for giving answers, the questions could be roughly classified in two types. The first type can be solved by extracting information from dialogues. Such type occupies 71.2$\%$ of all the questions. The second type can be replied not only by summarizing all the information from the given dialogues, but also need external medical knowledge and it occupies the rest 28.8$\%$. 

Since the questions are associated with medical knowledge and numerical data, we classify these questions based on how the data is used to give the answers. Three types of operations are generally used including: Judgement, Calculation and Inference. The following subsections introduce the details of each classified questions. 

\subsection{Judgement}
To answer the judgement questions, we usually need to find the corresponding evidences and statements. For instance, the question may ask does the patient need to take operation immediately. The second row of Figure 3 shows the Judgement questions. 

Generally, the type of questions can be solved by finding the statements from doctors in the dialogues. It is likely to obtain such statements at the end of the dialogues while the doctors give suggestions for the patients. However, most of the statements don’t directly answer the questions, for example, the doctor may say “You should stay at home and take some medicine.” instead of “You don’t need take operation.”, which means the reasoning ability is necessary to the final answer.

\begin{figure}[htb]
  \centering
  \includegraphics[scale=0.7]{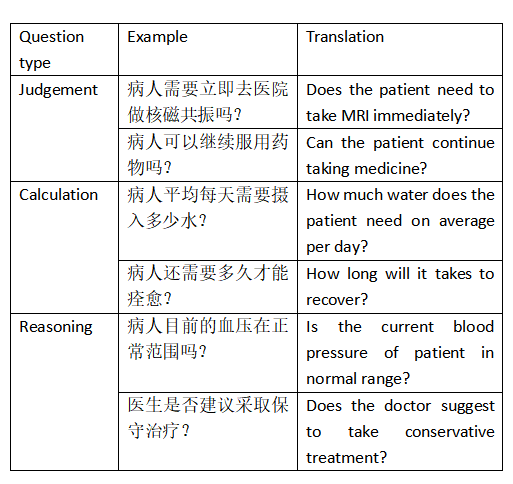}
  \caption{The examples of different types of questions} \label{fig3}
\end{figure}

\subsection{Calculation}
To solve the Calculation questions, the respondent should find and extract the corresponding figures from the dialogues and calculate the sum or difference. There are many different kinds of figures in the dialogues, including drug dose, the duration of disease, the date of diagnosis or operation, etc. 

Since all these figures are scattered, the respondent should look for the information correctly and efficiently, or the type and direction of the direction of calculation will be misjudged which leads to a wrong answer. Whatsmore, to give a correct answer, the respondent is required to have the ability of unit conversion, for example, a question may be “Can the patient recover in one week?”, while the doctor may say “The treatment will last at least half a month.”. A respondent should convert half a month into two weeks and compare it with “one week” in the question then get the answer. Similar to the Judgement questions, the Calculation questions still mainly depend on the corresponding text where the figures(evidence) are found. The third row in Table 3 shows some examples.

\subsection{Reasoning}
This type of questions requires the ability of combining information and reasoning. Generally, it needs a respondent to infer based on the collected information through the dialogue. 

For example, a question may be “Can the patient play basketball after taking the operation?”. Apparently, the doctor will not say “You can play basketball when recovered” or “You can’t play basketball anymore.”. The respondent may get the answer from “Your bones and joints will be completely recovered after the operation.”, which indicates that after the operation, the patient can do any kind of exercises including basketball. 

Besides, some of the questions not only require the ability to utilize the information obtained from the dialogues, but also the ability to use external medical ability. For instance, “Are the symptoms of the patient relevant to diabetes?”. To get the answer, a respondent should have the medical knowledge of diabetes which is not mentioned in the given dialogues. Therefore, this kind of questions becomes the most challenging part of the dataset.

\section{Experiments}
Based on the constructed dataset, we could define the QA task: the dataset
is comprised of triple instances \emph{(D, Q, A)} consisting of a dialogue \emph{D}, a natural language question \emph{Q} and a corresponding answer \emph{A}. \emph{D} is a set of several natural language sentences, the question \emph{Q} proposes a fact to be answered against the content in the dialogue \emph{D}, and the answer \emph{A} gives a response which could be a word,
a number, a phrase, or even a natural language sentence. During training, the model and the learning algorithm are presented with \emph{K} instances like $(D, Q, A)_{k=1}^K$ from the training set. In the testing stage, the model is presented with $(D, Q)_{k=1}^{K'}$ and supposed to predict the answer as \^{A}. The performance is measured by the prediction accuracy \emph{Acc} = $\frac{1}{K'}\sum_1^{K'} \left(\hat{A}_k = A_k\right)$ on the test set. We describe the proposed model and compared models as follows.

\subsection{MeDia-BERT}
In this approach, we view the dialogue QA task as a two-sequence binary classification problem since there are two candidate answers for each question. We put the dialogue \emph{D} into a sequence and treating the Question{Q} as another sequence. Due to the characteristics of dialogue compared with common text, to encode it as a sequence, we propose a hierarchical method.

Let $D = (u_1, u_2,..., u_T)$ denote a dialogue, where $C = (u_1, u_2,..., u_{T-1})$ is the dialogue context and $u_T$ is the response. Each $u_i = (w_1^i, w_2^i,...,w_{u_i}^i )$ in \emph{C} is an utterance and $w_k^i$ is the $k$-th word in $u_i$. In order to make the representation of \emph{C} better, we utilize a transformer (Vaswani et al. 2017) encoder with architecture. As shown in Figure 1, two transformer encoders are hierarchically nested: an utterance encoder to transform each utterance in \emph{C} to a vector and a context encoder to learn the representations of utterances and their neighbors in the context. We feed MeDiaQA questions and contexts, separated by the special \textmd{[SEP]} token, to pre-trained BERT (Devlin et al., 2019). The answers are predicted using the [CLS] token embedding using a softmax function. Cross-entropy loss of predicted and true answer distribution is denoted as \emph{L}. We finetune the model $\theta$ to minimize \emph{L} on the training set. 

\subsection{Compared Methods}
\noindent\textbf{Dominant:} Since more than 70$\%$ of questions are the kind of ‘yes or no’, where about 60$\%$ has the answer ‘no’, we choose ‘no’ as the default answer. And for the rest part, we choose the first option.

\noindent\textbf{BiLSTM:} We simply concatenate the dialogues and questions with learnable segment embeddings appended to the biomedical word2vec embeddings (Pyysalo et al., 2013) of each token. The concatenated sentence is then fed to a BiLSTM, and the final hidden states of the forward and backward network are used for giving the predicted answer.

\noindent\textbf{Human Performance:} To make a fairly evaluation of the difficulty of our dataset, we also hold out a small test set with 1K samples for human evaluation, where we distribute each (dialogue, question) pair to 3 different annotators to approximate human performance based on their majority voting.

\subsection{Experiments Results}
\begin{table}[h]
\caption{ The results of different baseline models on the test set.}\label{tab2}
\begin{center}
\begin{tabular}{|l|l|l|}
\hline \bf Model & \bf Accuracy (val) & \bf Accuracy (test)   \\ \hline
Dominant  & 56.6$\%$ & 57.1$\%$ \\
BiLSTM  & 54.2$\%$ & 55.1$\%$ \\
MeDiaBERT & 64.3$\%$ & 63.8$\%$ \\
Human Performance & 93.2$\%$ & 93.6$\%$ \\
\hline
\end{tabular}
\end{center}
\end{table}
The experiment results of four baseline models are shown with the accuracy on the test set in Table 4. The model which achieves the best performance is the proposed model MeDia-BERT, and the accuracy is 63.8$\%$ while the Dominant method achieves 57.1$\%$. However, the neural-network models, which used to perform well on other QA dataset, don’t perform better than the default answer method, and the BiLSTM achieves 55.1$\%$. And compared with human performance which achieves 93.6$\%$ on the test set, the performance of QA model seems relatively poor. In sum, the experiment results show that the MeDiaQA dataset is really challenging for the existing QA models.

\section{Conclusion}
In this paper, we construct MeDiaQA, a novel dataset for QA on medical dialogues and propose a method MeDia-BERT based on the state-of-the-art pre-trained language model. The proposed dataset shows the distinctiveness of medical dialogues compared with other normal domains in the context of QA. Due to the unique characteristics, it is challenging for existing QA models compared with human performance. We hope our dataset will lead a deeper research on machine reading comprehension in medical domain.

%
%
%
%

\end{document}